\documentclass[fleqn,10pt]{wlscirep}
\usepackage[utf8]{inputenc}
\usepackage[T1]{fontenc}
\newcommand{\ourmodel}[1]{MGL4MEP}
\usepackage{comment}
\usepackage{amsmath,amsfonts}
\usepackage{algorithmic}
\usepackage{algorithm}
\usepackage{array}
\usepackage[caption=false,font=normalsize,labelfont=sf,textfont=sf]{subfig}
\usepackage{textcomp}
\usepackage{stfloats}
\usepackage{url}
\usepackage{verbatim}
\usepackage{graphicx}
\usepackage{cite}
\usepackage{comment}
\usepackage{xcolor}
\usepackage{booktabs}
\usepackage{multirow}
\usepackage{makecell}
\usepackage{tabularx}

\usepackage{float}
\usepackage{hyperref}
\usepackage{bm}
\usepackage{amssymb}

\title{Multimodal Graph Learning for Modeling Emerging Pandemics with Big Data}

\author[1,2]{Khanh-Tung Tran}
\author[3]{Truong Son Hy}
\author[1]{Lili Jiang}
\author[1,*]{Xuan-Son Vu}
\affil[1]{Department of Computing Science, Ume\r{a} University, Sweden}
\affil[2]{AI Center, FPT Software, Hanoi, Vietnam}
\affil[3]{Department of Mathematics and Computer Science, Indiana State University, USA}

\affil[*]{Corresponding author. Email: sonvx@cs.umu.se}

\begin{abstract}
Accurate forecasting and analysis of emerging pandemics play a crucial role in effective public health management and decision-making. Traditional approaches primarily rely on epidemiological data, overlooking other valuable sources of information that could act as sensors or indicators of pandemic patterns. In this paper, we propose a novel framework called \ourmodel{} that integrates temporal graph neural networks and multi-modal data for learning and forecasting. We incorporate big data sources, including social media content, by utilizing specific pre-trained language models and discovering the underlying graph structure among users. This integration provides rich indicators of pandemic dynamics through learning with temporal graph neural networks. Extensive experiments demonstrate the effectiveness of our framework in pandemic forecasting and analysis, outperforming baseline methods across different areas, pandemic situations, and prediction horizons. The fusion of temporal graph learning and multi-modal data enables a comprehensive understanding of the pandemic landscape with less time lag, cheap cost, and more potential information indicators.
\end{abstract}
\begin{document}

\flushbottom
\maketitle
%
%
\thispagestyle{empty}

\section*{Introduction}

    Pandemics are global outbreaks of infectious diseases that affect many people across continents. The COVID-19 pandemic is one of the most significant pandemics of our time, impacting millions of individuals worldwide and causing lasting effects on our society. 
    In order to \emph{combat} pandemics, it is crucial to develop efficient solutions that facilitate the comprehension of their transmission and containment.
    This requires tracking and evaluating the evolution of pandemics through efficient monitoring and analysis of online resources that provide rich information, reflecting public knowledge and perceptions in a timely manner. For instance, 
    the volumes of social media interests can serve as early indicators of COVID-19 waves
    \cite{infodemics,9462520}, 
    and users' content can unveil diverse perspectives on regulations, such as quarantine measures or vaccination strategies.
    Understanding these signals can help policymakers to combat pandemics by recognizing trends in their spread and impact on the population, as well as the efficacy of current countermeasures.

    Traditional pandemic monitoring involves tracking hospital admissions, laboratory testing, and death rates, but can be expensive and lag in providing real-time disease spread updates. Compartmental models like SIR \cite{sir} and statistical analyses such as ARIMA \cite{arima} and Prophet \cite{prophet} use past data for predictions, are common approaches. However, these statistical models rely on assumptions and might lack data to precisely estimate factors like reproduction number in pandemic planning and forecasting.

    \begin{figure}[ht]
        \centering
        \includegraphics[width=0.6\columnwidth]{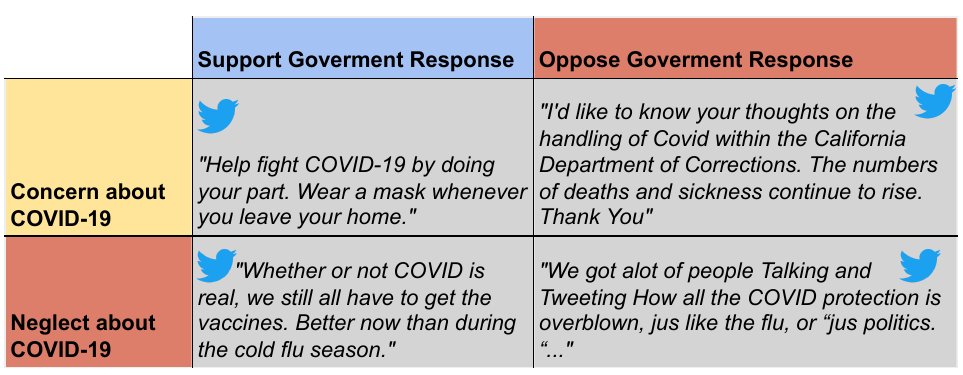}
        \vspace{-0.3cm}
        \caption{Examples showing different stances on social media reacting to the pandemic and  government regulations \cite{Tung2022}.}
        \label{fig1:motivation_ex}
    \end{figure}

    Time-series forecasting with deep learning is one of the most effective methods for tracking pandemics’ evolution. Because of their data-driven learning process, deep learning-based methods are highly accurate in analyzing time-series data to identify patterns and trends that can help predict future outbreaks through historical statistics. By using deep learning algorithms, we can analyze large amounts of data quickly and accurately, making it easier to identify patterns and trends that might be missed by other methods. Recent works leveraged deep learning-based methods to learn statistics from earlier time stamps as prediction to forecast the COVID-19 pandemic incidence, achieving better performance comparing to traditional methods \cite{Krba2020}.

    One of the limitations of previous works in pandemic forecasting is that they frequently rely entirely on epidemiological data and ignore other information that might act as sensors or indicators of the pandemic's patterns and evolution. Data from search engines, for example, can be used to monitor how individuals are looking for information about pandemics \cite{Dai2020,Higgins2020}. More crucially, social media data can be utilized to monitor how people are reacting to and feeling about a pandemic \cite{Tsao2021,Li2021}. Although previous research has examined the connection between social media usage and pandemic trends \cite{LAMSAL2022109603}, there has been little use of deep learning techniques to predict and track the spread of the epidemic. We can gain a more complete view of how an epidemic is evolving and how effective various treatments might be by including external knowledge from social media into pandemic forecasting models. For instance, by monitoring social media data, we may pinpoint public health campaigns to regions where people are most worried about a pandemic. Fig.~\ref{fig1:motivation_ex} illustrates our motivated examples, which show different stances of social media users on the COVID-19 pandemic and on government regulations. 
    
    During pandemics, social media has emerged as a key source of information, offering real-time updates on the spread of diseases and people's responses to them. Therefore, we investigate how pandemic tracking and analysis using deep learning algorithms can benefit from external knowledge from multi-modality, including social media and government regulations. More specifically, we construct graph-structured data from social media, treating each user as a node representing the current epidemic status. We dynamically capture interactions between users using temporal graph learning. Graph learning, particularly Graph Neural Networks (GNNs) \cite{gnn,ZHOU202057}, is an important branch of machine learning that deals with learning from and representing a variety of real-world relational data, including citation networks, social networks, knowledge graphs, etc. Incorporating graph learning techniques and graph-structured representations offers a promising approach to overcome limitations of previous works in pandemic forecasting, where graphs can capture the structural and semantic information of the pandemic domain \cite{panagopoulos2020transfer,keywordgraph1,keywordgraph2}.


    In this work, we introduce \ourmodel{}, a neural framework for forecasting and analyzing developing pandemics using big data sources and deep learning methods, including graph neural networks.
    We utilize the extremely recent COVID-19 pandemic and their effects on multiple areas as a case study. In order to trace and predict the evolution of the pandemic, we investigate the relationship between the pandemic risk factors and all other relevant data sources such as social media. Our framework will support many end users like politicians, policy makers, and general population for reference by providing complementary analysis and forecast information, leading to more effective crisis preventive and reaction times. Our contributions in this work are summarized as follows:

    \begin{itemize}
        \item We propose a multi-modal neural framework named \ourmodel{} for COVID-19 pandemic tracking and prediction,
        \item We extract and combine data from multiple sources, including social signals and government stringency signals as additional indicators to monitor pandemic trends and predict future evolution,
        \item We investigate the correlation and impacts of these multi-modal data on pandemic forecasting using deep learning and graph learning methods,
        \item We conduct extensive experiments on multiple areas affected by the pandemics to show the usefulness and effectiveness of our proposed framework.
    \end{itemize}

    Source codes of our framework and reproducible baselines are made publicly available at \url{https://github.com/KhanhTungTran/MGL4MEP} for future research and benchmarking purposes.

\section*{Results}
    \subsection*{Baselines}
        We evaluate our proposed approach against several baselines that employ different techniques, including statistical, machine learning, and deep learning approaches.
        \begin{itemize}
            \item Numerical analysis: (i) AVG: average of the whole history are used to predict the future; (ii) AVG\_WINDOW: average statistics of current prediction window are utilized to predict the future; and (iii) LAST DAY: the statistics of the current day are used as prediction.
            \item Machine learning-based models: (i) LIN\_REG: Ordinary least squares Linear Regression fits a line to training samples for predicting future cases; (ii) GP\_REG: Gaussian Process Regressor is a non-parametric regression model utilizes Gaussian processes; (iii) RAND\_FOREST and (iv) XGBOOST: tree-based models.
            \item Statistical models: (i) ARIMA \cite{arima}, a simple autoregressive moving average model leverage the entire history sequence as input; and (ii) PROPHET \cite{prophet}, similar to ARIMA but with strong seasonality characteristics.
            \item Deep learning models without graph topology: (i) A straightforward LSTM model, uses the sequence of the most recent $d$ days as its input, (ii) SE$_{transformer}$ and (iii) SRE$_{transformer}$, baseline models using the popular transformer architecture for learning on extracted text embeddings from social media data. Self-attention is calculated between tokens of different users (extracted using the same pre-trained language model as \ourmodel{} models). Then, the final embeddings are fused with LSTM for processing the time-series and making the final predictions. 
            \item \ourmodel{}$_{SR}$, \ourmodel{}$_{SE}$, \ourmodel{}$_{SRE}$: our proposed models (i) \ourmodel{}$_{SR}$ are baseline LSTM with an additional input features of regulations (R) information, (ii) \ourmodel{}$_{SE}$ is the model with temporal graph learning on pre-processed social media data (E - entity) from 1500 users for the default setting, and (iii) \ourmodel{}$_{SRE}$ is our final model with input from three different modality, including statistics as in traditional model, and regulations and social media data.
        \end{itemize}

    \subsection*{Implementation details}
        The proposed framework was implemented in PyTorch \cite{NEURIPS2019_bdbca288}, and experiments were carried out on an NVIDIA 3090Ti GPU. We train the model for a maximum of 300 epochs with early stopping. All models are optimized with AdamW optimizer \cite{loshchilov2018decoupled}, $10^{-3}$ initial learning rate, a batch size of 16, and input sequence length of 7. These hyperparameters are set empirically through grid search. All experiments are repeated 5 times with different seeds. The last 20\% time steps of the dataset are used as hold-out test set. Details regarding the data collected and used are described in following sections. \textit{Mean Absolute Error} (\textit{MAE}), \textit{Root Mean Squared Error} (\textit{RMSE}), \textit{Mean Absolute Percentage Error} (\textit{MAPE}), and \textit{R squared} ($R^2$) are metrics used to evaluate and compare between models.
        \begin{equation}
            \label{eq:mae}
            MAE = \frac{\sum_{i=1}^{n}|y_i-\hat{y}_i|}{n}
        \end{equation}
        where $y_i$ and $\hat{y}_i$ denote the $i$th statistics from ground truth data and predicted values of the models, and $n$ is the total number of samples in the test set. The MAE metric indicates the average variance between the predicted values and the ground truth in the dataset (lower is better).
        \begin{equation}
            \label{eq:rmse}
            RMSE = \sqrt{\frac{\sum_{i=1}^{n}(y_i-\hat{y}_i)^2}{n}}
        \end{equation}
        The RMSE metric in \autoref{eq:rmse} is the standard deviation of the residuals (prediction error) (lower is better).
        \begin{equation}
            \label{eq:mape}
            MAPE = \frac{1}{n}\sum_{i=1}^{n}\frac{|y_i-\hat{y}_i|}{y_i} \times 100
        \end{equation}
        The MAPE given in \autoref{eq:mape} tells us about the mean of the total percentage errors (lower is better).
        \begin{equation}
            \label{eq:r2}
            R^2 = 1 - \frac{RSS}{TSS}
        \end{equation}
        Finally, the Coefficient of Determination (R-squared metric) provides an insight into the similarity between real and predicted data, where the closer to 1 the R squared value is, the better. Here, $RSS=\sum_{i=1}^n(y_i-\hat{y}_i)^2$ and $TSS=\sum_{i=1}^n(y_i-\overline{y})^2$ denote the Residual Sum of Squares and the Total Sum of Squares, respectively.

        \begin{table*}[h]
        \begin{center}
        \caption{Results on California State for short-term predictions.}
        \label{tab:cali_short_term}
        \resizebox{1.0\linewidth}{!}{
        \begin{tabular}{l|cccc|cccc|cccc}
        \hline
        Number of days ahead & \multicolumn{4}{c|}{1} & \multicolumn{4}{c|}{3} & \multicolumn{4}{c}{7} \\
        \cline{2-13}
        & MAE$\downarrow$ & RMSE$\downarrow$ & MAPE$\downarrow$ & R$^2 \uparrow$ & MAE$\downarrow$ & RMSE$\downarrow$ & MAPE$\downarrow$ & R$^2 \uparrow$ & MAE$\downarrow$ & RMSE$\downarrow$ & MAPE$\downarrow$ & R$^2 \uparrow$ \\
        \Xhline{2\arrayrulewidth}
        AVG & 1735.92 & 2543.38 & 41.75 & -0.0729 & 1739.06 & 2552.49 & 42.13 & -0.0899 & 1726.68 & 2547.36 & 42.76 & -0.1460 \\
        LAST\_DAY & \textbf{200.24} & \underline{450.19} & \textbf{3.23} & \textbf{0.9820} & \underline{381.77} & \underline{730.72} & \textbf{6.44} & \textbf{0.9451} & \underline{638.60} & 1231.95 & 12.18 & \underline{0.8110} \\ 
        AVG\_WINDOW & 396.87 & 753.38 & 7.26 & 0.9352 & 527.41 & 990.43 & 10.08 & 0.8787 & 767.76 & 1337.56 & 15.76 & 0.7324 \\
        \hline
        
        LIN\_REG & 239.45 & 511.21 & 3.59 & 0.9761 & 465.00 & 853.38 & 6.61 & 0.9264 & 848.42 & 1649.43 & 13.13 & 0.6523 \\
        GP\_REG & 236.90 & 509.94 & 3.55 & 0.9762 & 457.30 & 843.30 & 6.50 & 0.9281 & 812.09 & 1585.71 & 12.53 & 0.6769\\
        RAND\_FOREST & 224.04 & 475.28 & 3.46 & 0.9785 & 504.18 & 906.01 & 7.44 & 0.9135 & 1251.66 & 2639.35 & 18.71 & 0.1089 \\
        XGBOOST & 265.20 & 581.56 & 4.46 & 0.9657 & 633.48 & 1166.90 & 10.35 & 0.8193 & 1511.12 & 3244.66 & 22.97 & -0.3955\\
        \hline
        ARIMA & 1536.31 & 2286.63 & 57.07 & -1.1243 & 1653.35 & 2431.21 & 60.55 & -1.3625 & 1874.52 & 2718.03 & 68.02 & -2.0135 \\
        PROPHET & 1919.07 & 3661.94 & 44.74 & -2.0139 & 1868.22 & 3618.93 & 44.44 & -1.9664 & 1732.04 & 3443.21 & 43.50 & -1.9019 \\
        LSTM & \underline{207.62} & \textbf{403.71} & \underline{4.41} & \underline{0.9780} & \textbf{349.35} & \textbf{663.60} & \textbf{6.46} & \underline{0.9395} & 831.11 & 1468.21 & 13.76 & 0.6564 \\
        SE$_{transformer}$ & 1026.56 & 1700.72 & 20.89 & 0.4242 & 975.09 & 1634.44 & 19.59 & 0.5352 & 1258.66 & 2136.51 & 23.93 & 0.1751 \\
        SRE$_{transformer}$ & 1437.27 & 2326.74 & 26.91 & 0.075 & 1193.21 & 1996.99 & 22.67 & 0.1598 & 1101.02 & 1845.40 & 21.61 & 0.3398 \\
        \Xhline{2\arrayrulewidth}
        \ourmodel{}$_{SR}$ (ours) & 265.77 & 470.65 & 5.16 & 0.9725 & 460.13& 783.48 & \underline{8.13} & 0.9217 & 825.72 & 1283.63 & 14.40 & 0.7568 \\
        \ourmodel{}$_{SE}$ (ours) & 510.36 & 854.95 & 9.73 & 0.9039 & 651.32 & 1081.28 & 12.59 & 0.8242 & \textbf{577.07} & \textbf{1005.08} & \textbf{11.07} & \textbf{0.8160} \\
        \ourmodel{}$_{SRE}$ (ours) & 479.11 & 788.15 & 9.41 & 0.9022 & 534.51 & 913.40 & 10.69 & 0.8755 & \underline{636.86} & \underline{1086.27} & \underline{12.08} & 0.7960\\
        \hline
        \end{tabular}
        }
        \end{center}
        \end{table*}

    \subsection*{Results}

        \begin{table*}[h]
        \begin{center}
        \caption{Results on New York State for short-term predictions.}
        \label{tab:ny_short_term}
        \resizebox{1.0\linewidth}{!}{
        \begin{tabular}{l|cccc|cccc|cccc}
        \hline
        Number of days ahead & \multicolumn{4}{c|}{1} & \multicolumn{4}{c|}{3} & \multicolumn{4}{c}{7} \\
        \cline{2-13}
        & MAE$\downarrow$ & RMSE$\downarrow$ & MAPE$\downarrow$ & R$^2 \uparrow$ & MAE$\downarrow$ & RMSE$\downarrow$ & MAPE$\downarrow$ & R$^2 \uparrow$ & MAE$\downarrow$ & RMSE$\downarrow$ & MAPE$\downarrow$ & R$^2 \uparrow$ \\
        \Xhline{2\arrayrulewidth}
        AVG & 549.85 & 842.34 & 32.40 & -3.8537 & 557.52 & 850.98 & 32.42 & -3.7875 & 566.90 & 861.35 & 33.01 & -3.9149 \\
        LAST\_DAY & 714.36 & 1243.29 & 21.06 & -0.5183 & 694.63 & 1146.65 & 22.49 & -0.6301 & 514.94 & 917.25 & 15.75 & \textbf{0.1593} \\ 
        AVG\_WINDOW & 493.52 & 824.75 & 15.68 & \underline{0.2165} & 505.44 & 839.23 & 16.18 & \textbf{0.1809} & 538.03  & 881.71 & 17.01 & \underline{0.1386} \\
        \hline
        LIN\_REG & 521.59 & 897.58 & 15.95 & 0.2221 & 485.24 & 806.84 & 15.54 & 0.2991 & 502.77 & 855.60 & 16.31 & 0.1978\\
        GP\_REG & 523.64 & 899.53 & 15.86 & 0.2199 & 486.08 & 807.60 & 15.40 & 0.3038 & 504.14 & 855.73 & 16.22 & 0.2073\\
        RAND\_FOREST & 1057.99 & 1996.79 & 29.76 & -2.0817 & 1134.78 & 2433.44 & 32.10 & -3.4917 & 607.52 & 1074.55 & 21.19 & -0.2863\\
        XGBOOST & 1307.55 & 3071.69 & 35.31 & -5.9887 & 1445.72 & 3411.18 & 40.63 & -7.7175 & 931.61 & 1587.23 & 31.59 & -1.7561 \\
        \hline
        ARIMA & 534.56 & 904.68 & 17.02 & -0.1033 & 543.39 & 912.94 & 17.13 & -0.0939 & 552.26 & 924.48 & 17.40 & -0.1036 \\
        PROPHET & 1565.38 & 2269.08 & 64.46 & -13.9960 & 1578.53 & 2339.01 & 63.30 & -14.6835 & 1582.25 & 2305.86 & 64.34 & -13.9820 \\
        LSTM & 186.66 & 301.33 & \underline{7.47} & 0.2098 & \underline{259.98} & \underline{417.70} & \underline{9.43} & -0.4364 & 366.56 & 593.31 & \underline{12.51} & -0.8850 \\
        SE$_{transformer}$ & 324.64 & 497.77 & 10.84 & -1.312 & 446.15 & 685.79 & 15.72 & -2.8800 & 377.39 & 573.80 & 13.36 & -1.4352 \\
        SRE$_{transformer}$ & 269.19 & 419.66 & 10.65 & -0.8072 & 339.15 & 543.18 & 11.72 & -1.3120 & \underline{353.17} & \underline{556.10} & \textbf{11.47} & -1.4280 \\
        \Xhline{2\arrayrulewidth}
        \ourmodel{}$_{SR}$ (ours) & 203.35 & 320.73 & 7.77 & -0.0129 & 386.10 & 575.82 & 11.22 & -0.9012 & 462.88 & 696.32 & 13.02 & -1.6530 \\
        \ourmodel{}$_{SE}$ (ours) & \textbf{160.63} & \textbf{278.49} & \textbf{5.97} & \textbf{0.2927} & \textbf{242.02} & \textbf{409.45} & \textbf{8.46} & -0.1808 & \textbf{336.77} & \textbf{549.74} & \textbf{11.40} & -0.7996 \\
        \ourmodel{}$_{SRE}$ (ours) & \underline{179.52} & \underline{285.52} & 7.70 & 0.1352 & 294.04 & 461.59 & 9.91 & -0.5289 & 565.34 & 836.36 & 14.77 & -2.5251 \\
        \hline
        \end{tabular}
        }
        \end{center}
        \end{table*}

        The evaluation results for short-term predictions are presented in Table~\ref{tab:cali_short_term}, where it can be observed that for horizon values of 1 or 3, our proposed approaches underperform the baseline LSTM neural network method, which is the closest competitor to our proposed models. However, the performance of our models significantly improve on the New York state dataset, as shown in Table~\ref{tab:ny_short_term}. The reason for this improvement may lie in the fact that the $R^2$ score, which is calculated on the predictions for each state, being close to perfection ($0.9820$ for the horizon of 1) on the California dataset, but the $R^2$ score on New York dataset is significantly lower, indicating a higher level of difficulty for learning and prediction on the latter dataset. In general, the \ourmodel{}$_{SE}$ method achieves the most impressive results for short-term forecasting, followed by the \ourmodel{}$_{SRE}$ model. This can be explained as the time lag between government efforts and their impact on the real-world situation spans multiple weeks. Incorporating this information for short-term prediction may adversely affect the effectiveness of the model. Furthermore, upon comparison with SE$_{transformer}$ and SRE$_{transformer}$, which utilize the transformer architecture without correlation matrices for processing social media data, we observe that our methodology incorporating graph neural networks featuring spatial-temporal characteristics distinctively surpass these approaches. This outcome highlights the efficacy and suitability of our novel proposed approaches, both in constructing input graph structures and in learning algorithms, for addressing these types of multi-modality domains.

        \begin{table*}[!htp]
        \begin{center}
        \caption{Results on California State for long-term predictions.}
        \label{tab:cali_long_term}
        \resizebox{1.0\linewidth}{!}{
        \begin{tabular}{l|cccc|cccc|cccc}
        \hline
        Number of days ahead & \multicolumn{4}{c|}{14} & \multicolumn{4}{c|}{21} & \multicolumn{4}{c}{28} \\
        \cline{2-13}
        & MAE$\downarrow$ & RMSE$\downarrow$ & MAPE$\downarrow$ & R$^2 \uparrow$ & MAE$\downarrow$ & RMSE$\downarrow$ & MAPE$\downarrow$ & R$^2 \uparrow$ & MAE$\downarrow$ & RMSE$\downarrow$ & MAPE$\downarrow$ & R$^2 \uparrow$ \\
        \Xhline{2\arrayrulewidth}
        AVG & 1625.33 & 2397.80 & 43.58 & -0.4115 & 1550.94 & 2299.96 & 44.92 & -1.2001 & 1620.44 & 2367.44 & 47.82 & -3.4092 \\
        LAST\_DAY & 1016.07 & 1679.24 & 22.26 & 0.3751 & 1454.70 & 2418.62 & 34.59 & -1.2151 & 1854.50 & 3011.52 & 47.58 & -5.9595 \\ 
        AVG\_WINDOW & 1190.22 & 1943.52 & 26.61 & 0.1444 & 1582.72 & 2578.15 & 38.55 & -1.6106 & 1988.98 & 3196.71 & 52.41 & -6.9129 \\
        \hline
        LIN\_REG & 1073.60 & 1882.31 & 20.53 & 0.0937 & 1393.40 & 2469.55 & 28.95 & -2.1887 & 1668.01 & 2917.30 & 36.16 & -7.5977 \\
        GP\_REG & 1004.46 & 1740.18 & 19.67 & 0.1889 & 1357.33 & 2433.31 & 28.26 & -2.1518 & 1629.51 & 2837.48 & 35.48 & -7.5078 \\
        RAND\_FOREST & 2800.75 & 6404.53 & 44.94 & -8.4948 & 5310.35 & 11285.29 & 96.23 & -61.9137 & 6253.21 & 11610.95 & 123.79 & -119.0820 \\
        XGBOOST & 2905.03 & 6253.72 & 51.53 & -9.3011 & 6129.52 & 12290.41 & 112.19 & -70.4298 & 6426.62 & 11945.64 & 123.59 & -119.2901\\
        \hline
        ARIMA & 2249.92 & 3203.26 & 81.65 & -4.2380 & 2660.01 & 3697.61 & 96.55 & -9.7618 & 3057.48 & 4167.76 & 111.46 & -23.8195 \\
        PROPHET & 1417.82 & 2917.99 & 41.58 & -2.0017 & 1081.00 & 2227.87 & 39.39 & -2.4306 & 848.07 & 1726.27 & 37.21 & -3.4281 \\
        LSTM & 962.75 & 1464.46 & 21.91 & 0.3687 & 855.98 & 1324.40 & 20.05 & -0.2023 & 981.55 & 1557.89 & 22.00 & -1.2480 \\
        SE$_{transformer}$ & 1250.80 & 2248.76 & 23.58 & -0.2571 & 950.71 & 1792.47 & 20.24 & -0.3664 & 879.30 & 1532.81 & 18.36 & -1.0662 \\
        SRE$_{transformer}$ & 1044.82 & 1938.24 & 20.73 & 0.0852 & 1136.17 & 1988.57 & 21.95 & -0.5690 & 658.05 & 1199.98 & 16.37 & -0.3351 \\
        \Xhline{2\arrayrulewidth}
        \ourmodel{}$_{SR}$ (ours) & 771.67 & 1236.55 & \underline{17.04} & 0.5435 & \underline{836.22} & \underline{1285.58} & \underline{17.24} & \underline{0.1384} & 866.02 & 1440.28 & 18.65 & -0.4867 \\
        \ourmodel{}$_{SE}$ (ours) & \underline{583.17} & \underline{1035.97} & \textbf{12.10} & \underline{0.6798} & \textbf{563.19} & \textbf{932.89} & \textbf{13.08} & \textbf{0.4467} & \textbf{588.17} & \textbf{1036.76} & \textbf{13.55} & \textbf{0.1767} \\
        \ourmodel{}$_{SRE}$ (ours) & \textbf{555.80} & \textbf{981.07} & \textbf{12.15} & \textbf{0.6866} & \textbf{568.03} & \textbf{948.07} & \textbf{12.85} & \textbf{0.4457} & \underline{634.88} & \underline{1149.32} & \underline{14.52} & \underline{-0.1925} \\
        \hline
        \end{tabular}
        }
        \end{center}
        \end{table*}
        With respect to long-term prediction, our models achieve the best results across all three horizons with significant gaps compared to all other methods, as illustrated in Table~\ref{tab:cali_long_term} and \ref{tab:ny_long_term}. Generally, the best performed approaches are \ourmodel{}$_{SRE}$ and \ourmodel{}$_{SE}$ models, achieving 42.47\%, 34.21\%, and 10.62\% lower MAE with horizon equals 14, 21, 28 days ahead than the best basline methods for California dataset, and 11.94\%, and 7.50\% lower MAE for horizon 14 and 21, on New York dataset. Moreover, for long-term prediction, \ourmodel{}$_{SR}$ models are able to obtain better results than the simple baseline LSTM models, compared to the results and analysis on short-term predictions. The model's capability on forecasting the long-term trajectory of the pandemic means that it can provide valuable information and insights for government and policy makers on planning ahead and making informed, timely response to the pandemic.

        \begin{table*}[!htp]
        \begin{center}
        \caption{Results on New York State for long-term predictions.}
        \label{tab:ny_long_term}
        \resizebox{1.0\linewidth}{!}{
        \begin{tabular}{l|cccc|cccc|cccc}
        \hline
        Number of days ahead & \multicolumn{4}{c|}{14} & \multicolumn{4}{c|}{21} & \multicolumn{4}{c}{28} \\
        \cline{2-13}
        & MAE$\downarrow$ & RMSE$\downarrow$ & MAPE$\downarrow$ & R$^2 \uparrow$ & MAE$\downarrow$ & RMSE$\downarrow$ & MAPE$\downarrow$ & R$^2 \uparrow$ & MAE$\downarrow$ & RMSE$\downarrow$ & MAPE$\downarrow$ & R$^2 \uparrow$ \\
        \Xhline{2\arrayrulewidth}
        AVG & 576.57 & 878.21 & 36.03 & -7.4231 & 577.12 & 878.64 & 37.71 & -9.4983 & 555.38 & 854.57 & 38.84 & -9.9887 \\
        LAST\_DAY & 608.37 & 1096.63 & 19.75 & -0.5469 & 568.43 & 1041.60 & 21.58 & -0.9110 & 649.36 & 1128.66 & 25.78 & -1.2729 \\ 
        AVG\_WINDOW & 577.85 & 950.80 & 19.97 & -0.3342 & 561.28 & 918.91 & 21.37 & -0.6601 & 576.23 & 941.92 & 23.50 & -0.9414 \\
        \hline
        LIN\_REG & 550.95 & 901.68 & 20.44 & -0.5076 & 542.15 & 917.15 & 23.91 & -1.5887 & 577.39 & 984.80 & 29.15 & -3.5869\\
        GP\_REG & 550.54 & 900.15 & 20.24 & -0.4716 & 539.68 & 913.79 & 23.59 & -1.5046 & 575.04 & 979.01 & 28.76 & -3.4214\\
        RAND\_FOREST & 962.95 & 1649.01 & 34.08 & -3.4251 & 1267.22 & 2077.25 & 42.03 & -5.2131 & 1339.23 & 2165.87 & 51.17 & -10.8534\\
        XGBOOST & 1242.85 & 2078.66 & 49.98 & -10.5987 & 1753.07 & 2790.31 & 65.61 & -15.9673 & 1688.39 & 2611.17 & 87.04 & -63.6253\\
        \hline
        ARIMA & 554.70 & 927.95 & 22.08 & -0.8887 & 546.87 & 918.12 & 23.62 & -1.3794 & 529.60 & 902.33 & 25.27 & -1.6065 \\
        PROPHET & 1579.78 & 2307.45 & 64.20 & -13.8898 & 1575.09 & 2298.21 & 63.94 & -15.1467 & 1584.90 & 2306.69 & 63.77 & -15.9274 \\
        LSTM & 270.61 & 434.96 & 11.08 & -0.6286 & 288.95 & 452.27 & 10.12 & -0.4052 & \textbf{280.53} & \textbf{443.59} & 11.53 & -1.022 \\
        SE$_{transformer}$ & 393.33 & 596.95 & 12.67 & -1.2141 & 318.92 & 506.33 & 11.41 & -0.8533 & 324.25 & 511.64 & 11.89 & -1.1520 \\
        SRE$_{transformer}$ & 393.19 & 643.84 & 12.71 & -1.6740 & 366.67 & 560.50 & 13.42 & -1.5570 & 368.12 & 564.89 & 11.52 & -0.7978 \\
        \Xhline{2\arrayrulewidth}
        \ourmodel{}$_{SR}$ (ours) & \underline{267.15} & \underline{430.25} & \underline{10.42} & -0.5040 & \textbf{254.03} & \textbf{411.54} & \textbf{9.74} & \textbf{-0.3239} & 315.59 & 504.57 & \underline{10.86} & \underline{-0.7395} \\
        \ourmodel{}$_{SE}$ (ours) &  \textbf{238.29} & \textbf{382.46} & \textbf{9.47} & \textbf{-0.2762} & \underline{267.29} & \underline{431.71} & \underline{10.04} & \underline{-0.3655} & \underline{311.72} & \underline{503.53} & 11.29 & -0.8625 \\
        \ourmodel{}$_{SRE}$ (ours) & 278.71 & 460.88 & \underline{10.44} & \underline{-0.3482} & 287.07 & 479.69 & \textbf{9.73} & -0.3946 & 324.18 & 520.57 & \textbf{10.56} & \textbf{-0.6624} \\
        \hline
        \end{tabular}
        }
        \end{center}
        \end{table*}

        We conduct comprehensive ablation studies to investigate the impact of the size of the input social media graph on our COVID-19 prediction model. In particular, we reduce the number of users selected for building the graph from the original 1,500 users to 1,000 and 500, respectively. Our hypothesis is that as we decrease the number of users, the amount of information provided by the social media data would be significantly reduced. The results in Table~\ref{tab:cali_num_nodes} for the California region confirm our hypothesis, showing a degradation in performance with a decrease in the number of nodes for both short-term and long-term forecasting. The findings have implications for future research in that it is critical to take the size of the input social media network into account when developing a model for predicting COVID-19 instances. The size of the social media graph directly influences the richness and diversity of the data captured, allowing the model to capture a more nuanced understanding of public sentiments, behaviors, and trends related to the pandemic. 

        \begin{table*}[!htp]
        \begin{center}
        \caption{Ablation study on sufficient amount of social media users for modeling society interaction and factor. We train \ourmodel{}$_{SE}$ with different number of nodes (users) for social network data and evaluate each model's performances on California data.}
        \label{tab:cali_num_nodes}
        \resizebox{0.85\linewidth}{!}{
        \begin{tabular}{l|cccc|cccc}
        \hline
        Number of days ahead & \multicolumn{4}{c|}{7} & \multicolumn{4}{c}{14} \\
        \cline{2-9}
        & MAE$\downarrow$ & RMSE$\downarrow$ & MAPE$\downarrow$ & R$^2 \uparrow$ & MAE$\downarrow$ & RMSE$\downarrow$ & MAPE$\downarrow$ & R$^2 \uparrow$ \\
        \hline
        LSTM - without social network info. & 831.11 & 1468.21 & 13.76 & 0.6564 & 962.75 & 1464.46 & 21.91 & 0.3687 \\
        \Xhline{2\arrayrulewidth}
        \ourmodel{}$_{SE}$ - 500 nodes & 687.43 & 1139.88 & 13.74 & 0.7783 & 630.78 & 1049.50 & 13.75 & 0.6631 \\
        \hline
        \ourmodel{}$_{SE}$ - 1000 nodes & 626.24 & 1047.17 & 12.74 & 0.7918 & 595.77 & 1051.18 & 12.74 & 0.6636 \\
        \hline
        \ourmodel{}$_{SE}$ - 1500 nodes & \textbf{577.07} & \textbf{1005.08} & \textbf{11.07} & \textbf{0.8160} & \textbf{583.17} & \textbf{1035.97} & \textbf{12.10} & \textbf{0.6798} \\
        \hline
        \end{tabular}
        }
        \end{center}
        \end{table*}

        Table~\ref{tab:ny_num_nodes} illustrates the ablation study on different number of nodes for the input social media graph of our COVID-19 prediction model for New York dataset. The results are consistent with experiment on California dataset, where a degradation in performance with a decrease in the number of nodes for both short-term and long-term forecasting can be seen.

        \begin{table*}[ht]
                \begin{center}
                \caption{Ablation study on sufficient amount of social media users for modeling society interaction and factor. We train our \ourmodel{}$_{SE}$ model with different number of nodes (users) for social network graph and evaluate each model's performances on New York data.}
                \label{tab:ny_num_nodes}
                \resizebox{0.85\linewidth}{!}{
                \begin{tabular}{l|cccc|cccc}
                \hline
                \multirow{2}{*}{Number of days ahead} & \multicolumn{4}{c|}{7} & \multicolumn{4}{c}{14} \\
                \cline{2-9}
                & MAE$\downarrow$ & RMSE$\downarrow$ & MAPE$\downarrow$ & R$^2 \uparrow$ & MAE$\downarrow$ & RMSE$\downarrow$ & MAPE$\downarrow$ & R$^2 \uparrow$ \\
                \hline
                LSTM - without social network info. & 366.56 & 593.31 & 12.51 & -0.8850 & 270.61 & 434.96 & 11.08 & -0.6286 \\
                \Xhline{2\arrayrulewidth}
                \ourmodel{}$_{SE}$ - 500 nodes & 371.54 & 596.19 & 12.40 & -1.3730 & 240.97 & 395.27 & \textbf{9.49} & -0.3487 \\
                \hline
                \ourmodel{}$_{SE}$ - 1000 nodes & 346.40 & 570.16 & 12.65 & -1.5580 & 249.31 & 405.01 & 9.68 & -0.3383 \\
                \hline
                \ourmodel{}$_{SE}$ - 1500 nodes & \textbf{336.77} & \textbf{549.74} & \textbf{11.40} & \textbf{-0.7996} & \textbf{238.29} & \textbf{382.46} & \textbf{9.47} & \textbf{-0.2762} \\
                \hline
                \end{tabular}
                }
                \end{center}
            \end{table*}

\begin{figure*}[!htp]
        \centering
        \subfloat[]{\includegraphics[width=2.8in]{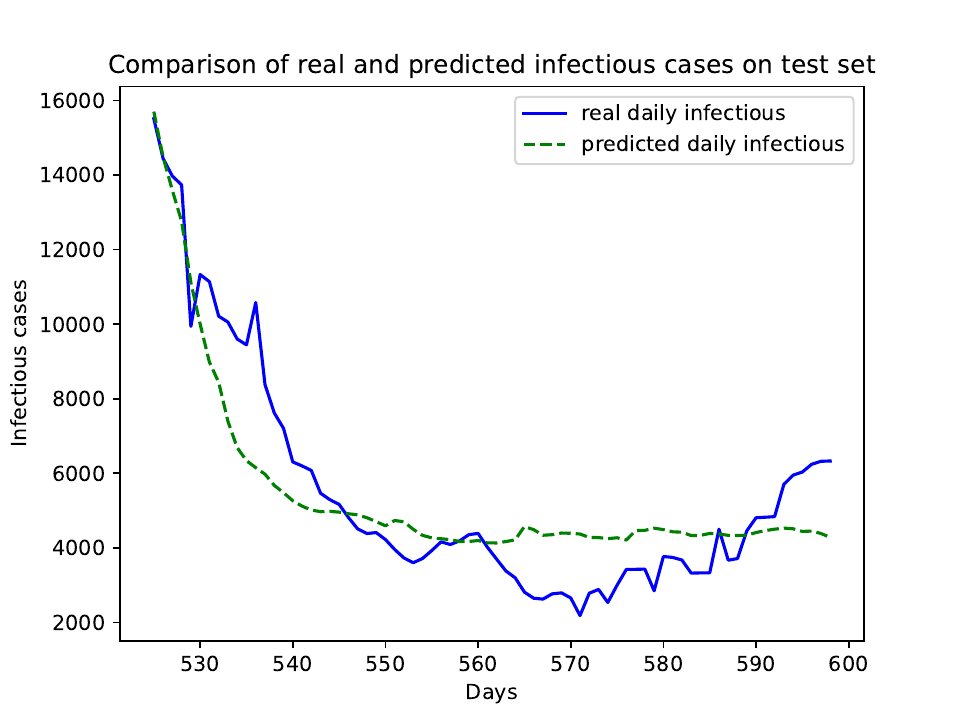}%
        }
        \hfil
        \subfloat[]{\includegraphics[width=2.8in]{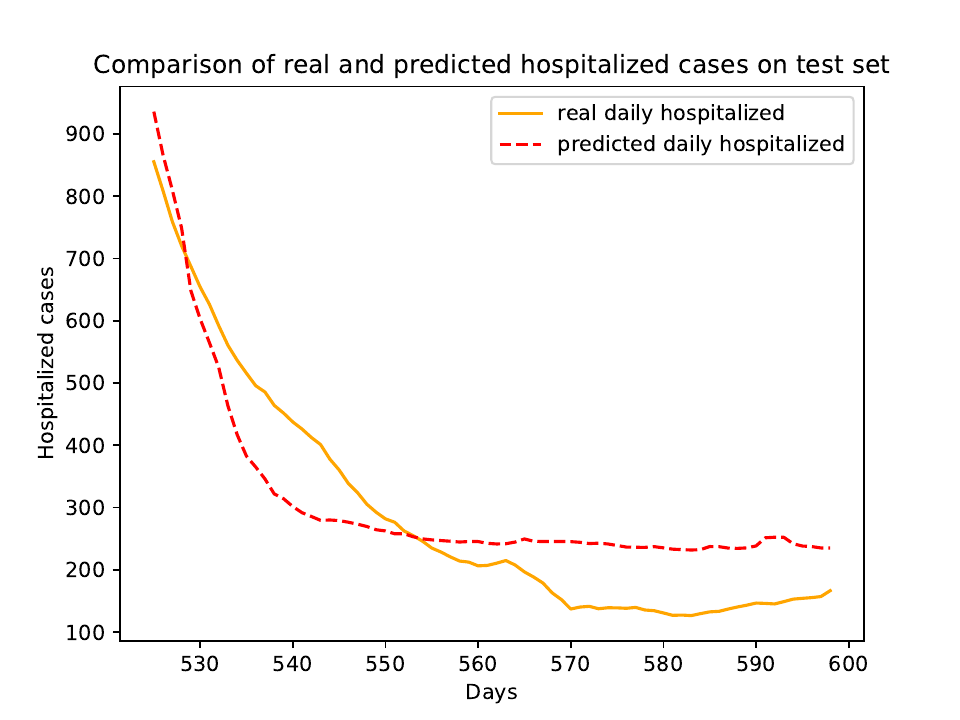}%
        }
        \caption{Forecasting results on test set for the newly collected period of California state dataset (a) Infectious cases. (b) Hospitalized cases.}
        \label{fig:prediction_cali_expand}
        \end{figure*}

        In order to assess the usefulness and effectiveness of our proposed methods in different stages of the pandemic, we perform an additional experiment by collecting data for another 150 days, thereby increasing the total amount of data collected. We then train and evaluate the same models and baselines on this new dataset, the results of which are presented in Table~\ref{tab:cali_longer} and visualized in Fig.~\ref{fig:prediction_cali_expand}. It is worth noting that this new test set for California state exhibits a higher variance compared to the previous forecasting range, where statistics gradually decrease with sharp changes. Nevertheless, our models outperform the baselines and achieve the best performance on both test sets. These results indicate the robustness and generalizability of our approaches in combating the COVID-19 pandemic in different stages of its evolution.

        \begin{table*}[ht]
        \begin{center}
        \caption{Results on California State when the data collection period went up to May, 2022.}
        \label{tab:cali_longer}
        \resizebox{0.77\linewidth}{!}{
        \begin{tabular}{l|cccc|cccc}
        \hline
        Number of days ahead & \multicolumn{4}{c|}{7} & \multicolumn{4}{c}{14} \\
        \cline{2-9}
        & MAE$\downarrow$ & RMSE$\downarrow$ & MAPE$\downarrow$ & R$^2 \uparrow$ & MAE$\downarrow$ & RMSE$\downarrow$ & MAPE$\downarrow$ & R$^2 \uparrow$ \\
        \Xhline{2\arrayrulewidth}
        AVG & 5409.71 & 8863.60 & 194.54 & -0.2504 & 4563.77 & 6692.37 & 206.27 & -1.0961 \\
        LAST\_DAY & 3357.62 & 9021.44 & 34.76 & 0.3118 & 5453.62 & 13683.49 & 76.84 & -3.5250 \\ 
        AVG\_WINDOW & 5179.47 & 13189.69 & 52.49 & -0.4659 & 7428.87 & 18219.47 & 104.02 & -6.8496\\
        \hline
        LIN\_REG & 3930.85 & 9618.96 & 20.27 & 0.9333 & 5852.20 & 14373.83 & 38.48 & 0.7726\\
        GP\_REG & 3942.71 & 9627.01 & 20.22 & 0.9331 & 5858.40 & 14406.63 & 38.23 & 0.7701 \\
        RAND\_FOREST & 8364.92 & 22422.34 & 28.10 & 0.6728 & 7599.95 & 19360.56 & 45.53 & 0.6717 \\
        XGBOOST & 8051.32 & 21248.67 & 27.10 & 0.7062 & 7646.05 & 18724.51 & 48.16 & 0.6739 \\
        \hline
        ARIMA & 9568.85 & 17111.98 & 146.64 & -1.1930 & 11804.89 & 21882.42 & 203.03 & -9.3336 \\
        PROPHET & 35891.74 & 52742.08 & 1009.45 & -20.6104 & 36453.48 & 52415.34 & 1051.09 & -57.9110 \\
        LSTM & 1384.07 & 2801.17 & 29.70 & 0.7748 & 932.91 & 1691.75 & \textbf{32.58} & \underline{0.5906} \\
        SE$_{transformer}$ & 1549.09 & 3709.52 & 56.29 & 0.4954 & 1096.73 & 2052.00 & 45.93 & 0.2553 \\
        SRE$_{transformer}$ & 1844.22 & 5042.73 & 50.15 & 0.1378 & 1285.99 & 2359.96 & 50.56 & 0.0341 \\
        \Xhline{2\arrayrulewidth}
        \ourmodel{}$_{SR}$ (ours) & 1099.96 & 2303.54 & \textbf{28.36} & \underline{0.7907} & 1124.82 & 2744.64 & 38.89 & -0.0915 \\
        \ourmodel{}$_{SE}$ (ours) & \underline{1057.29} & \underline{2085.78} & 30.51 & \textbf{0.7972} & \textbf{734.44} & \textbf{1250.40} & \underline{37.20} & \textbf{0.6592} \\
        \ourmodel{}$_{SRE}$ (ours) & \textbf{901.99} & \textbf{1881.81} & \underline{28.95} & 0.7818 & \underline{828.84} & \underline{1375.84} & 37.45 & 0.5682 \\
        \hline  
        \end{tabular}
        }
        \end{center}
        \end{table*}

\section*{Discussion}
    \textbf{Forecasting results.}\quad In the real world, COVID-19 is a complex pandemic, and many factors that cannot be seen by previous statistics can lead to different future scenarios. For example, if a government implements a strict lockdown, the number of new cases will likely decrease. However, if a government lifts all restrictions, the number of new cases will likely increase. This is why it is important to consider multiple information sources when forecasting the spread of COVID-19. Our proposed method, \ourmodel{} and its variants incorporate multiple information sources effectively, leading to better performances, lower errors, and sustained accuracy, particularly in long-term predictions, compared to other popular forecasting models. \ourmodel{} enjoys these benefits due to it being able to learn the dynamic relationships between various factors that affect the spread of COVID-19, such as official government policy, and social stances against the pandemic or situation. Moreover, our ablation results clearly demonstrate that the availability of more information significantly enhances the reliability of our forecasting models. Comparisons between our methods and baselines that do not utilize the graph structure of social media data also highlight the efficacy of our graph-structure generation process and temporal graph learning framework. The results not only underline the effectiveness in learning and extracting information of the models, but also emphasize on the usefulness of input multimodal data.
    
    Additionally, our experiment results show that \ourmodel{} is adaptable and robust to different situations and history. This means that it can be used to forecast the spread of COVID-19 in different countries and regions, even if the pandemic is evolving rapidly. The predictions made by \ourmodel{} can be leverage by different beneficial group such as the authorities to develop appropriate strategies in order to deal with the spread of this pandemic. An example is that \ourmodel{} can be used to forecast the impact of different government policies on the spread of the virus. 
    
    Finally, it's important to highlight the automated nature of the forecasting process with \ourmodel{}. The entire process can be automated and seamlessly updated whenever new information becomes available. This automation is possible due to our framework's reliance on openly accessible data from the Internet, which can be efficiently gathered through automated web crawling.

    \textbf{Model limitations.}\quad Although some of the proposed models performed well according to certain metrics, we found several shortcomings in the models that we tested. One limitation is that \ourmodel{} takes time to realize the trend depending on the dynamic at the considered area, as information from different sources, such as the effects of government policies takes time to be reflected in the data. For example, in the case of New York state, it takes effect immediately while in the case of California, it takes about 14 days. Additionally, the underlying factors that affect the infection of COVID-19 are diverse, and it can be difficult to capture all of them through the multiple data sources used by \ourmodel{}. Another limitation of \ourmodel{} is that similar to other deep learning methods, it is a black-box model. This means that we cannot easily understand how it makes its predictions, and can make it difficult to trust or explain the model's predictions.

    \textbf{Future research directions.}\quad There are several ways in which \ourmodel{} can be improved in the future. An interesting future research direction is to enrich our framework with more information regarding the pandemic situation, such as regional age population, mobility, or virus variants. Another direction can be interested in explainability methods, such as identifying important nodes or features through temporal graph learning, or understanding the most valuable factors that affect the forecasting results. This would make us more confident in the predictions of the model and would help us to better understand the dynamics of the pandemic.

\section*{Related Work}
    \subsection*{Pandemic forecasting}
        Traditional approaches leverage statistical models have been widely used to forecast COVID-19. These approaches involve analyzing past epidemic data using statistical and time-series methodologies to identify patterns and trends, which can then be used to forecast future outbreaks. 
        Methods like autoregressive integrated moving average (ARIMA) \cite{arima} and Prophet \cite{prophet} are effective at identifying trends in stationary time-series data and handling periodic patterns, respectively. One alternative to forecast pandemics, such as COVID-19 pandemic, is through the use of compartmental models \cite{Postnikov2020,FernndezVillaverde2022,Tang2020}. These models divide a population into compartments, such as susceptible, exposed, and infected individuals (SIR model), and use mathematical equations to describe the dynamic and transitions between each group. However, it is important to acknowledge that leveraging statistical models for pandemic forecasting has its limitations, as they presumptively assume a linear relationship between past and future time-series. Such methods rely on certain assumptions and may lack the data necessary to accurately address all the relevant issues. 

        Deep learning have been applied to make predictions about the spread of COVID-19 pandemic and achieve high performances \cite{Liu2020,Krba2020}. With the enormous dataset of records such as infected and hospitalized cases collected on a daily basis, deep learning is considered a suitable approach, as neural networks can learn and update from data effectively. 
        Sequential models such as Recurrent Neural Network (RNN) and Long Short Term Memory (LSTM) \cite{lstm} have been applied and seen high performances for forecasting COVID-19 pandemic, both at world-wide or country level \cite{9378276,Chandra2022}, and more fine-grained levels including state and county levels \cite{Nikparvar2021,Lucas2022,pmlr-v184-hy22a, Panagopoulos_Nikolentzos_Vazirgiannis_2021}. Unlike conventional approaches, deep learning can incorporate external knowledge and adapt to changing circumstances, improving their predictive capabilities. These approaches, or fusion of multiple neural network models, can incorporate a wide range of data sources, including social media, to provide a more comprehensive view of the pandemic and its potential impacts. 
        
        To our best knowledge, previous research has only attempted to integrate basic indicators or indices, overlooking the dynamic, intricate information contained in user-generated content \cite{Dai2020,10.3389/fpubh.2022.806813,LAMSAL2022109603}. 
        Incorporating these valuable signals into neural networks remains a challenge but has the potential to provide a more comprehensive view of the pandemic and its potential impacts \cite{Higgins2020,Ibrahim2021,infodemics}.

    \subsection*{Leveraging external resources for time-series forecasting}
        Previous studies have explored the connection between social media interests and pandemic trends. In \cite{Higgins2020}, the authors highlight a strong correlation between peak of search volume on COVID-19 pandemic and the development of the pandemic, upto 20 days earlier than the issuance of official warnings. The authors of \cite{infodemics} also discovered a close connection between the evolution of the COVID-19 crisis and social media user's sentiments toward different phases of the pandemic. Another work \cite{Bae2021} makes use of the social impact of media coverage to support the compartment model for pandemic prediction. Post-processed indicators such as internal movement index and economic response have been incorporated as additional input features to sequence models for forecasting future statistics \cite{Khan2021,Xu2022}. 
        Differing from them, our method considers every aspect of user response through social media and government regulations against the pandemic. To achieve greater accuracy in pandemic forecasting, we analyze individual tweets and search for relevant social events.

        There has been a significant amount of interest in effectively leveraging social media as an external knowledge source for more accurate pandemic forecasting. In \cite{10.3389/fpubh.2022.806813}, the authors used tweet count (the amount of tweets related to COVID-19) per day as an additional input to an LSTM model and achieve better results than using statistics only. 
        Taking a step further, in \cite{LAMSAL2022109603}, the collected tweets are then further extracted into two main features, representing user sentiment and topic of interest. These features are used as additional input features to an ARIMAX model, which is an extension of ARIMA.
        Furthermore, in \cite{CHEW2021107417}, important keywords are extracted and curated into a keyword cloud to present the most important information for each day and input to a MLP module for pandemic prediction. Perhaps the most relevant works to this paper are \cite{keywordgraph1,keywordgraph2} where the authors extract the most popular keywords per day and view them as a graph structure and employ graph algorithms to learn on those representations. 
        
        In this study, in contrast to prior works, we incorporate data from multiple different sources, with social media as an important knowledge source where we build graph structure with each user as node, or an indicator on the current status of the epidemic, and dynamically represent the interaction between them through temporal graph neural networks \cite{gnn,gcrn,Wu2019}. Our approach comprehensively considers various aspects of user responses on social media and government regulations pertaining to the pandemic.


    \subsection*{Temporal graph neural networks forecasting models}
        Graph neural networks (GNNs) have gained significant attention in various learning tasks, such as image recognition \cite{chen2019multi,Nguyen2021}, estimating quantum chemical computation \cite{duvenaud2015convolutional,10.5555/3305381.3305512,Hy2018}, predicting protein interfaces \cite{NIPS2017_f5077839}, etc. GNNs generalize the concept of convolution neural networks to non-Euclidean domains, allowing for local operations on the nodes and edges of a graph \cite{gnn,defferrard2016convolutional}. The most popular GNNs is Message Passing Neural Networks (MPNNs) \cite{10.5555/3305381.3305512} in which the graph convolution is defined via the message passing scheme that propagates and then aggregates the vectorized information between each node and its local neighborhood.

        To handle evolving features and connectivity over time, temporal graph neural networks have been introduced. Unlike static graphs, temporal graphs are usually represented by a sequence of node interactions over continuous time instead of an adjacency matrix. Temporal GNNs aim to capture both the temporal and structural information of the temporal graphs by introducing a node memory that represents the state of the node at a given time, acting as a compressed representation of the node’s past interactions. Temporal GNNs combine graph encoding techniques with time-series encoding architectures such as LSTM and Transformers, forming a powerful deep learning framework. 
        They find applications in various domains, such as traffic prediction, where they outperform traditional methods by incorporating spatial relationships of road networks and temporal dynamics of traffic conditions \cite{li2018diffusion,agcrn,nguyen2023fast}. In the analysis of brain networks, temporal GNNs utilize invasive techniques like electrocorticography (ECoG) to uncover temporal patterns and gain insights into brain network dynamics \cite{nguyen2023fast}. 

        In our apporach, by leveraging the temporal and structural aspects of graph representation in social media data, temporal GNNs enhance modeling capabilities for understanding evolving complex systems and forecasting pandemic statistics.

\section*{Preliminaries}
    \subsection*{Time-series forecasting}
        Originally proposed in \cite{lstm}, Long short-term memory (LSTM) has been the dominant recurrent network architecture for learning from sequences of data. Unlike standard feedforward neural networks, LSTM can process and retain the temporal correlations between adjacent time steps, due to its feedback connections. For a historical time step $t$, the output $y_t$ will not only depend on $x_t$ but also from previous iterations through hidden state $h_{t-1}$ and memory variable $c_{t-1}$:
        \begin{subequations}
        \label{eq:lstm}
        \begin{gather}
            \Gamma_u = \sigma (\bm{W}_{hu}h_{t-1} + \bm{W}_{xu}x_{t} + b_{u}) \\
            \Gamma_f = \sigma (\bm{W}_{hf}h_{t-1} + \bm{W}_{xf}x_{t} + b_{f}) \\
            \Tilde{c}_t = \tanh(\bm{W}_{hc}h_{t-1} + \bm{W}_{xc}x_t + b_c) \\
            c_t = \Gamma_u \odot \Tilde{c}_t + \Gamma_f \odot c_{t-1} \\
            \Gamma_o = \sigma (\bm{W}_{ho}h_{t-1} + \bm{W}_{xo}x_{t} + b_{o}) \\
            h_t = \Gamma_o \odot \tanh(c_t)
        \end{gather}
        \end{subequations}
        where $\Gamma_u$ and $\Gamma_f$ are ``update gate'' and ``output gate'', calculated through a sigmoid ($\sigma$) activation function to determine the percentage of new memory $\Tilde{c}_t$ to keep and the percentage of old memory $c_{t-1}$ to forget, respectively. The ``output gate'' $\Gamma_o$ allows information to be revealed appropriately due to the sigmoid function then the weights are updated by the element-wise multiplication of $\Gamma_o$ and memory cell $c_t$ activated by the non-linear $\tanh$ function.

        A simpler, more intuitive version of LSTM called Gated-Recurrent Unit (GRU) \cite{gru}, combined the cell memory and the hidden state variable into $h_t$ to transfer information. Therefore, a GRU only has two gates, a ``reset gate'' and an ``update gate''.
        \begin{subequations}
        \label{eq:gru}
        \begin{gather}
            \Gamma_u = \sigma (\bm{W}_{hu}h_{t-1} + \bm{W}_{xu}x_{t} + b_{u}) \\
            \Gamma_f = 1 - \Gamma_u \\
            \Tilde{h}_t = \tanh(\bm{W}_{hh}h_{t-1} + \bm{W}_{xh}x_t + b_h)    \\
            h_t = \Gamma_u \odot \Tilde{h}_t + \Gamma_f \odot h_{t-1}
        \end{gather}
        \end{subequations}

        Finally, the last hidden state variable $h_t$ can be used to predict the corresponding output value $\hat{y}_t$ through a fully connected layer with $softmax$ activation function:
        \begin{equation}
            \hat{y}_t = softmax(\bm{W}_{hy} h_t)
        \end{equation}

    \subsection*{Temporal graph learning algorithms}
        Graph neural networks (GNNs) are a class of neural networks that operate on graph-structured data. Graphs are a powerful method to represent many types of data, such as social networks, biological networks, and traffic flows. 
        GNNs are capable of learning the relationships between nodes in a graph.
        They generalize the concept of convolutional neural networks to non-Euclidean domains by defining local operations on the nodes and edges of a graph. A typical GNN layer operate on input graph $\mathcal{G}=(\bm{X},\bm{E},\bm{A})$ can be formulate as in \autoref{eq:GNN}.
        \begin{equation}
            \label{eq:GNN}
            \bm{Y} = g_{\bm{W}} \star \bm{X} \approx (\bm{I_N} + \bm{D}^{-\frac{1}{2}}\bm{A}\bm{D}^{-\frac{1}{2}})\bm{X}\bm{W}
        \end{equation}
        where $\bm{X} \in \mathbb{R}^{N\times D_X}$ represents node matrix, each of the $N$ nodes has $D_X$ features, and $\bm{A} \in \mathbb{R}^{N\times N}$ is a weighted adjacency matrix encoding set of edges $\bm{E}$. The graph convolution operator $\star$ can be approximated by first-order Chebyshev polynomial expansian and generalized to high-dimensional \cite{defferrard2016convolutional,kipf2017semisupervised} with learnable parameter $\bm{W} \in \mathbb{R}^{D_X \times D_Y}$.

        Temporal Graph Neural Networks are an extension of GNNs that can handle temporal graphs, i.e., graphs that change over time. Unlike static graphs, temporal graphs are usually represented by a sequence of node interactions over continuous time instead of an adjacency matrix. Temporal GNNs aim to capture both the temporal and structural information of the temporal graphs by introducing a node memory that represents the state of the node at a given time, acting as a compressed representation of the node’s past interactions. 
        In this work, we follow the framework of recent studies, including GCRN \cite{gcrn}, AGCRN \cite{agcrn}, and MPNN LSTM \cite{panagopoulos2020transfer}, that utilize recurrent neural network on top of graph convolution operators. We leverage a simplified approach and use GRU as the recursive network architecture.
        \begin{subequations}
        \label{eq:tgn}
        \begin{gather}
            \bm{\Gamma}_U = \sigma (g_{\bm{W}_{HU}} \star \bm{H}_{t-1} + g_{\bm{W}_{XU}} \star \bm{X}_{t} + \bm{b_{U}}) \\
            \bm{\Gamma}_F = 1 - \bm{\Gamma_U} \\
            \bm{\Tilde{H}}_t = \tanh(g_{\bm{W}_{HH}} \star \bm{H}_{t-1} + g_{\bm{W}_{XH}} \star \bm{X}_t + \bm{b_H})    \\
            \bm{H}_t = \bm{\Gamma}_U \odot \bm{\Tilde{H}}_t + \bm{\Gamma}_F \odot \bm{H}_{t-1}
        \end{gather}
        \end{subequations}

        This framework allows \ourmodel{} to learn the dynamic interactions between entities, act as nodes, or indicators to the current status of the pandemic and between different time stamps throughout the evolution of the pandemic.

    \subsection*{Pre-trained Language models}
        \label{sec:pretrain_lang}
        Since we are dealing with free-text data to capture population's reactions to the pandemic, more specifically, user-generated content sourced from social media, it is crucial to extract meaningful information before constructing the graph-structured representation of the data. 
        In recent years, large pre-trained language models have revolutionized the field of natural language processing such as Bidirectional Encoder Representations from Transformers (BERT) \cite{Devlin2019}. These models have demonstrated exceptional capabilities in understanding and generating human language. BERT's underlying architecture, based on Transformer \cite{10.5555/3295222.3295349}, employs self-attention mechanisms to capture dependencies between words or tokens in a sentence. This enables BERT to comprehend the contextual information of a word based on its surrounding words, leading to more accurate language understanding and representation. By pre-training BERT on massive amounts of textual data and a wide variety of tasks, such as masked language modeling and next sentence prediction, the model learns a rich language representation that can be fine-tuned for specific downstream tasks.
        
        Building upon recent advancements in applying large pre-trained models to domain-specific data \cite{10.1093/bioinformatics/btz682,10.1145/3458754}, we leverage BertTweet \cite{bertweet}, a variant of BERT as our main feature extractor for text embeddings. The model has been trained on a large amount of Twitter data, especially including a sub-set of COVID-19 related data, and its effectiveness in capturing the nuanced meanings and signals conveyed in text data has been well-established. By leveraging BertTweet, we can obtain high-quality features that accurately represent the semantic content of social media posts surrounding the COVID-19 pandemic, and enables us to discover valuable patterns and trends that contribute to a comprehensive understanding of the social media landscape during the global health crisis.

\section*{Methodology}

    \subsection*{\ourmodel{}}
        \begin{figure*}[ht!]
            \centering
            \includegraphics[width=1.0\textwidth]{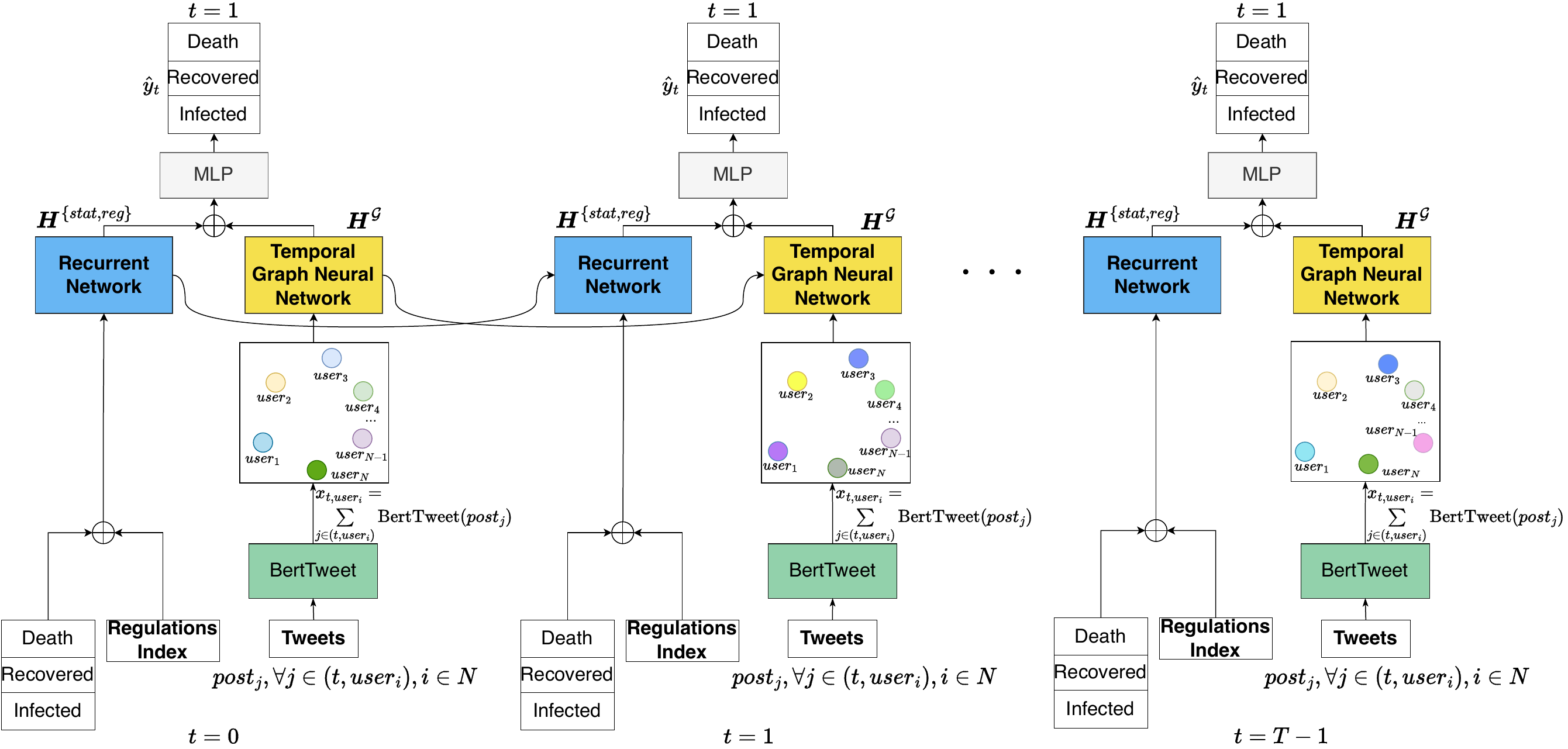}
            \caption{Overall architecture of 
            \ourmodel{} - a multi-modal framework for enhanced pandemic forecasting with external resources. \ourmodel{} incorporates both pandemic related metrics and population's reactions on social media into the forecasting to better capture the dynamic properties of emerging pandemics.}
            \label{fig:multimodal_framework}
        \end{figure*}

        \begin{figure*}[ht!]
            \centering
            \subfloat[]{\includegraphics[width=3.1in]{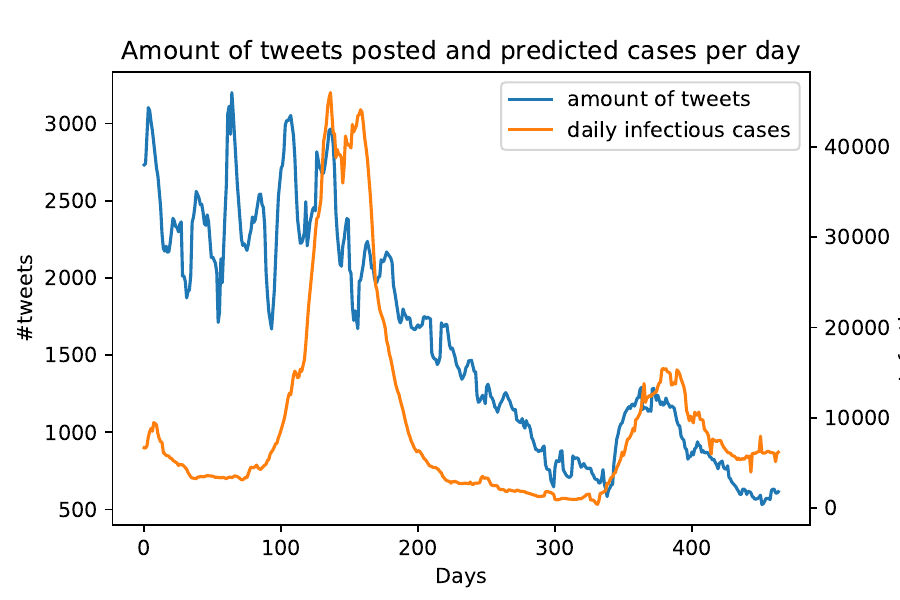}%
            \label{fig_first_case}}
            \hfil
            \subfloat[]{\includegraphics[width=3.1in]{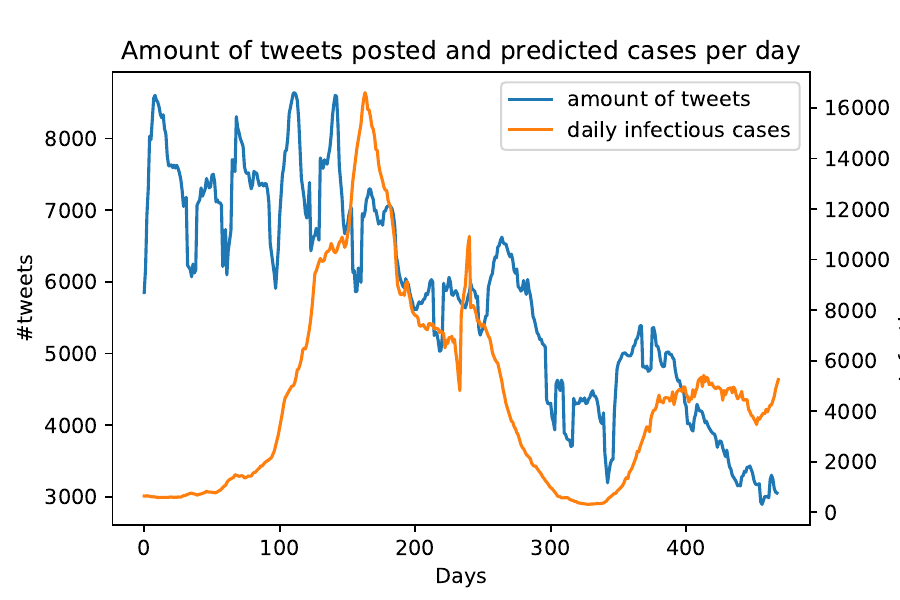}%
            \label{fig_second_case}}
            \caption{Comparison between amount of tweets posted and number of new COVID-19 cases per day. (a) in California state. (b) in New York state.}
            \label{fig:tweet_corr}
        \end{figure*}

        In this section, we present the multi-modality framework and techniques employed in our study to effectively extract and model multi-modal data for COVID-19 forecasting. Our approach aims to harness the power of social media data, specifically user-generated content, and government stringency index, to gain valuable insights to the evolution of the pandemic. We describe the key components of our framework, including data pre-processing, feature extraction, graph-based representation with temporal graph learning, and multi-modal learning. The main components of our proposed framework are depicted in \autoref{fig:multimodal_framework}.

        One goal of our multi-modality framework is to effectively incorporate signals from human-generated text data through social media platform, offering valuable reflections of the population's response to the pandemic, as exampled in \autoref{fig1:motivation_ex}.  This importance can also be underscored by the sheer amount of Covid-19 discussion over time, strongly correlating with pandemic statistics, as shown in \autoref{fig:tweet_corr}.
        Details about our social media data collection process is provided in the next Section. Extracting meaningful features from text data is crucial for constructing a comprehensive understanding of the information shared on social media platforms. 
        While other works proposed another direction of extracting indices like sentiment, this might discard necessary information, such as sentiment about COVID-19, but not about current government response to the pandemic. 
        As discussed in previous sections, by employing pre-trained language model, specifically BertTweet, as text feature extractor, we ensure to capture rich insights contained within user-generated content, especially in relation to COVID-19 pandemic. Our framework enables integrating various modalities to capture the complex and temporal dynamics of emerging pandemics. We obtain temporal embedding for each user by utilizing BertTweet on their text data as follows:
        \begin{equation}
            x_{t, user_i} = \sum_{j \in (t, user_i)}\text{BertTweet}(post_j),
        \end{equation}
        where $t$ denotes the timestamp, $user_i$ denotes the i-th user in our social media data, and $post_j$, $j \in (t, user_i)$ denotes the text obtained from the tweets of the i-th user at time $t$. The inclusion of user interactions and shared information across users is crucial for further analysis. In order to capture the correlations and dependencies between users, it is imperative to construct a graph-structured representation of the pre-processed data. This graph-based approach allows us to model the interactions and information flow through graph neural networks, capturing the dynamics and valuable insights related to the ongoing pandemic from social media signals. Hence, we introduce an end-to-end learning algorithm to discover the underlying graph structure that captures the correlation among time-series in a data-driven manner. More specifically, define node embeddings extracted from pre-trained language model as  $\bm{X}^{\mathcal{G}}_t := [x_{t, user_1}, x_{t, user_2}, \ldots, x_{t, user_N}]^T \in \mathbb{R}^{N\times D_X},$ and the continuous adjacency matrix can be calculated as the dot-product similarity matrix of the node embeddings: $\bm{A}^\mathcal{G}_t = \bm{X}^{\mathcal{G}}_t \cdot {\bm{X}^{\mathcal{G}}_t}^T \in \mathbb{R}^{N \times N}$.
        However, to enable effective learning with temporal graph learning algorithms, there are two downsides with this approach: first, large embedding dimension lead to incorrect adjacency matrix calculation. Second, may include information not directly related to our downstream task, and take up resources in training and evaluating. Inspired from AGCRN \cite{agcrn}, we employed a node-specific learnable embeddings that allows us to map input dimension to a lower intermediate embedding dimension:
        \begin{equation}
            g_{\bm{W}, \bm{E}} \star \bm{X} = (\bm{I_N} + \text{softmax}(\text{ReLU}(\bm{E} \cdot \bm{E}^T)) \bm{X} \bm{E} \bm{W},
        \end{equation}
        where $g$ denotes the filter parameterized by W and E, while $\star$ denotes the graph convolution operator. $\bm{E}$ is a learnable intermediate node embedding matrix, $\bm{E} \in \mathbb{R}^{N \times D_{emb}}$. The input node matrix is multiplied with the node embedding $\bm{E}$, resulting in an updated representation, where $\bm{E}$ is learnable, meanings that the representation is specific for each node and its pattern. Then, the integrated node embeddings are further multiplied by the weight matrix $\bm{W}$ to incorporate the influence of the node-specific features. Moreover, we replace the normalized graph Laplician matrix \cite{kipf2017semisupervised} by computing the inner product of the intermediate node embedding matrix $\bm{E}$ with its transpose $\bm{E}^T$. This operation captures the pairwise relationships between node embeddings and produces a matrix of shape $N \times N$. We apply the rectified linear unit (ReLU) activation function to introduce non-linearity and ensure positive values in the resulting matrix. The $\text{softmax}$ function is then applied to normalize the values across matrix row, ensuring that the row sums to 1. This step allows us to obtain a valid probability distribution representing the importance or relevance of each node, or each user, with respect to others.
    
        This graph convolution operation is plugged into the framework in \autoref{eq:tgn}, and the final temporal graph learning algorithm is shown in \autoref{eq:final_tgn}:
        \begin{subequations}
        \label{eq:final_tgn}
        \begin{gather}
            \bm{\Gamma}_U^\mathcal{G} = \sigma (g_{\bm{W}_{HU}, \bm{E}_H} \star \bm{H}_{t-1}^\mathcal{G} + g_{\bm{W}_{XU}, \bm{E}_X} \star \bm{X}_{t}^\mathcal{G} + \bm{b}_{U}) \\
            \bm{\Gamma}_F^\mathcal{G} = 1 - \bm{\Gamma_U}^\mathcal{G} \\
            \bm{\Tilde{H}}_t^\mathcal{G} = \tanh(g_{\bm{W}_{HH}, \bm{E}_H} \star \bm{H}_{t-1}^\mathcal{G} + g_{\bm{W}_{XH}, \bm{E}_X} \star \bm{X}_t^\mathcal{G} + \bm{b}_H)    \\
            \bm{H}_t^\mathcal{G} = \bm{\Gamma}_U^\mathcal{G} \odot \bm{\Tilde{H}}_t^\mathcal{G} + \bm{\Gamma}_F^\mathcal{G} \odot \bm{H}_{t-1}^\mathcal{G}
        \end{gather}
        \end{subequations}
        where $g$ denotes learnable weights with respect to different embeddings.
        To complement the multi-modal nature of our framework, we incorporate government stringency features that provide valuable insights into the pandemic response at a regional level. Government stringency features capture the level of restrictions, policies, and interventions implemented by authorities to mitigate the spread of COVID-19. These features serve as an important contextual signal to enhance the understanding of the evolving dynamics in our model.
    
        Specifically, we utilize the raw data and formula proposed in \cite{covid_gov_response} to compute an indicator on the level of government stringency. However, recognizing the complexity of this domain, we compare and analyze each individual indicator, as well as the averaged general stringency index, to identify the most suitable indicator for the current pandemic situation. 
        In \autoref{fig:string_corr}, we present the correlation levels between index and the number of new COVID cases for two indicators, with different time lags. The results suggest a strong relationship between the restriction on internal movement to the status of the pandemic. Hence, in the refined version of our framework, we leverage this specific indicator as a measure of government stringency. 

        Since this indicator can be represented as a vector for each day, similar to the statistical metrics of the pandemic, we can employ a sophisticated recurrent neural network (i.e., \autoref{eq:lstm} and \autoref{eq:gru}) to learn solely on this feature. Alternatively, we have the option to combine, or concatenate it with the pandemic statistics and learn through a unified recurrent network. Through extensive experimentation, we have found that the latter approach yields superior performance, and thus, it is our final choice for incorporating the government stringency indicator into our framework.
    
        Finally, in order make accurate predictions, it is crucial to integrate the information from multiple modalities in our framework. We achieve this by fusing the embeddings obtained from different modalities, namely statistical features, government stringency features, and social-media graph-based features. The fusion process is performed using the equation:
        \begin{equation}
            \hat{y}_{t+T} = \text{softmax}(\bm{W} (\bm{H}_{t+T}^{\{stat, reg\}} \oplus \bm{H}_{t+T}^\mathcal{G}))
        \end{equation}
        where $\bm{H}_{t+T}^{\{stat, reg\}}$ is the learned embeddings of recurrent neural network for statistical metrics, $\hat{y}_{t+T}$ represents the predicted value for the time step $t+T$, where $T$ is the forecasting horizon. Embeddings from various domain, capturing the relevant information for each modality, are fused using the concatenation operator $\oplus$ to create a unified feature representation. 
    
        Using the aforementioned equation to integrate the embeddings from multi-modality, our system successfully combines a variety of information sources while utilizing the complementary nature of different modalities for enhanced forecasting performance. This comprehensive approach enables us to capture the intricate dynamics and interdependencies within the data, leading to more accurate and reliable forecasts for the future evolution of the pandemic.

    \subsection*{Multimodal data collection process}
        In this study, as shown in Table~\ref{tab:input_features}, we utilized three different types of data sources to gain insights into the COVID-19 pandemic and its development.

        \begin{table}[h]
        \caption{Input features used for proposed approaches}
        \begin{center}
        \begin{tabularx}{0.7\linewidth}{X|X}
        \hline
        \textbf{Data Source} & \textbf{Features}\\
        \hline
        JHU CSSE COVID-19 Data \cite{Dong2020} & Daily COVID-19 Statistics  \\
        \hline
        Oxford Covid-19 Government Response Tracker \cite{covid_gov_response} & Government Stringency Index \\
        & Rate of Change of Stringency Index over a time period \\
        & Restrictions on Internal Movement Indicator \\
        & Rate of Change of Restrictions on Internal Movement Indicator over a time period \\
        \hline
        Twitter \cite{epidemiologia2030024} & Daily user-generated contents from users with topics-of-interest related to COVID-19 \\
        \hline
        \end{tabularx}
        \label{tab:input_features}
        \end{center}
        \end{table}
    
\begin{figure}[ht!]
        \centering
          \includegraphics[width=0.6\linewidth]{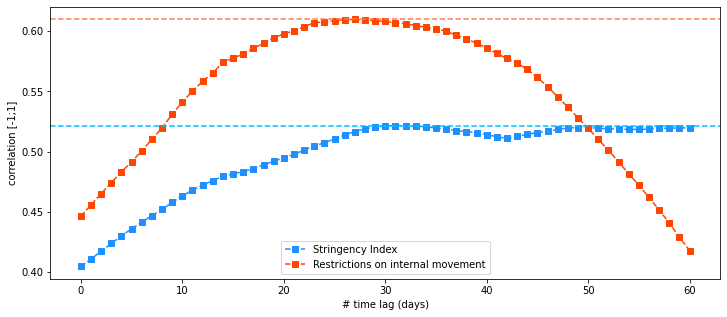}
          \caption{Correlations between Stringency Index or Restrictions on internal movement to daily infected cases of COVID-19 across different time-lags.}
         \label{fig:string_corr}
        \end{figure}

        \begin{itemize}
            \item \textbf{COVID-19 Statistic Data.} We leverage the statistic dataset from Johns Hopkins University \cite{Dong2020} with 450 data points from August 1, 2020, to November 30, 2021. Each data point is represented as the number of confirmed COVID-19 infections or serious, hospitalized cases in a given area per day. Our final task is time-series forecasting on this multi-variate statistics with different horizons to predict the trajectory of the pandemic. Then, trained models can be a valuable tool in responding to the pandemic, as it can support policymakers give better decisions about how to allocate resources, implement public health measures, and prepare for the future.
            \item \textbf{COVID-19 Government Responses and Regulations Data.} The stringency index data \cite{covid_gov_response} is a valuable resource in understanding the level of government response to the COVID-19 pandemic. The index is represented as a numeric value between $0$ to $100$ and includes nine different indicators, such as the closure of schools and workplaces, cancellation of public events, restrictions on gatherings, and orders to shelter in place. Fig.~\ref{fig:string_corr} displays the correlation values between the stringency index and record restrictions on internal movement between regions and the daily statistics of new infected cases. Interestingly, both time-lag horizons exhibit a clear trend of correlation values peaking at around 30 days. Moreover, the correlation of record restrictions on internal movement with new infected cases is consistently higher than that of the stringency index. This is also observed when considering new hospitalized cases. The found correlation trends imply that the current government response can act as a valuable indication to forecast how the epidemic will develop in the future.
            \item \textbf{Social Media Data.} 
            We crawl a total amount of more than 74 million tweets using Twitter API and tweets ids of all tweets related to COVID-19 released by Banda et al. \cite{epidemiologia2030024}. The original authors leveraged Twitter Stream for collecting all tweets in the category of COVID-19 chatter, with over 4 million tweets a day. We filtered out tweets with geo-location tags in either California state or New York state in this exploratory study. Moreover, we filtered out all tweets that are not in English. We randomly keep all tweets from 1,500 different users for each location. The distributions of tweets over time with respect to statistics of newly confirmed cases are illustrated in Fig.~\ref{fig:tweet_corr}. A strong correlation between the two time-series can be recognized, although there is a noisy period at the start. This is likely due to the initial confusion and fear surrounding the appearance of COVID-19, which led to a high volume of discussions about the virus worldwide. As the situation became more stable and people gained a better understanding of the pandemic affection on their own regions, the amount of tweets posted became more relevant and had a higher correlation with users' areas of residence. To account for this, we excluded data from the initial few months of the pandemic and only collected data starting from August 1, 2020.
        \end{itemize}

\section*{Conclusion}
    In this work, we present a novel framework named \ourmodel{} that combines temporal graph neural networks and multi-modal data for accurate pandemic forecasting. By integrating various big data sources, including social media content, we effectively capture the complex dynamics of emerging pandemics. Our framework outperforms traditional approaches by leveraging the potential of pre-trained language models and generating graph-structured data. Extensive experiments conducted with multiple variants of our proposed method demonstrate the effectiveness of our framework in providing timely and comprehensive insights into the pandemic landscape. The fusion of temporal graph learning and multi-modal data enables a deeper understanding of the evolving patterns and indicators, leading to more informed public health management and decision-making. Our approach offers a promising direction for leveraging big data in pandemic research and provides a foundation for future advancements in the field.

\bibliography{main}

\section*{Acknowledgements}

The project is partially funded by the STINT Mobility Grants for Internationalisation in Sweden (no: MG2020-8848). The computations was enabled by the supercomputing resource Berzelius provided by National Supercomputer Centre at Linköping University and the Knut and Alice Wallenberg foundation.

\section*{Author contributions}
    Conceptualization: T.S.H., L.J., and X.S.V. Methodology: K.T.T., L.J., and X.S.V. Software: K.T.T. Analysis: K.T.T., T.S.H. Validation: L.J., and X.S.V. Paper preparation: K.T.T., T.S.H., L.J., and X.S.V. All authors reviewed the manuscript.

\section*{Data availability}
    The COVID-19 statistics leveraged in this study are available from Johns Hopkins University Center for Systems Science and Engineering in the COVID-19 repository, \url{https://github.com/CSSEGISandData/COVID-19}. The COVID-19 Government Responses and Regulations Data is sourced from The Oxford Covid-19 Government Response Tracker (OxCGRT), \url{https://github.com/OxCGRT/covid-policy-dataset}. The social media data are crawled from Twitter social media platform by tweets related to COVID-19 released by Covid-19 Twitter chatter dataset for scientific use, \url{https://github.com/thepanacealab/COVID19_twitter}.

\section*{Competing interests}
    The authors declare no competing interests.

\end{document}